\documentclass[Afour,sageh,times]{sagej}

\usepackage{moreverb,url}

\usepackage{upgreek}
\usepackage{amsmath} 
\usepackage{amssymb}
\newcommand{\diag}{\mathop{\mathrm{diag}}}
\usepackage{esvect}
\usepackage{multirow}
\usepackage{makecell}

\usepackage[colorlinks,bookmarksopen,bookmarksnumbered,citecolor=red,urlcolor=red]{hyperref}

\newcommand\BibTeX{{\rmfamily B\kern-.05em \textsc{i\kern-.025em b}\kern-.08em
T\kern-.1667em\lower.7ex\hbox{E}\kern-.125emX}}

\def\volumeyear{2024}

\begin{document}

\runninghead{HWIL for Embedded State Estimation in Microrobots}

\title{Hardware-in-the-Loop for Characterization of Embedded State Estimation for Flying Microrobots}

\author{Aryan Naveen\affilnum{1}, Jalil Morris\affilnum{1}, Christian Chan\affilnum{1}, Daniel Mhrous\affilnum{1}, E. Farrell Helbling\affilnum{2}, Nak-Seung
Patrick Hyun\affilnum{3}, Gage Hills\affilnum{1}, and Robert J. Wood\affilnum{1}}

\affiliation{\affilnum{1}Harvard University School of Engineering and Applied Sciences, Cambridge, MA \\
\affilnum{2}Cornell University School of Electrical and Computer Engineering, Ithaca, NY \\
\affilnum{3}Purdue University School of Electrical and Computer Engineering, West Lafayette, IN}

\corrauth{Aryan Naveen, 
150 Western Ave, Boston, MA 02134}

\email{aryan\_naveen@college.harvard.edu}

\begin{abstract}
Autonomous flapping-wing micro-aerial vehicles (FWMAV) have a host of potential applications such as environmental monitoring, artificial pollination, and search and rescue operations. One of the challenges for achieving these applications is the implementation of an onboard sensor suite due to the small size and limited payload capacity of FWMAVs. The current solution for accurate state estimation is the use of offboard motion capture cameras, thus restricting vehicle operation to a special flight arena. In addition, the small payload capacity and highly non-linear oscillating dynamics of FWMAVs makes state estimation using onboard sensors challenging due to limited compute power and sensor noise. In this paper, we develop a novel hardware-in-the-loop (HWIL) testing pipeline that recreates flight trajectories of the Harvard RoboBee, a 100mg FWMAV. We apply this testing pipeline to evaluate a potential suite of sensors for robust altitude and attitude estimation by implementing and characterizing a Complimentary Extended Kalman Filter. The HWIL system includes a mechanical noise generator, such that both trajectories and oscillatinos can be emulated and evaluated. Our onboard sensing package works towards the future goal of enabling fully autonomous control for micro-aerial vehicles.
\end{abstract}


\maketitle
\section{Introduction}
\label{sec: intro}

Autonomous flapping-wing micro-aerial vehicles (FWMAV) possess the unique combination of small size and high maneuverability, opening a wide range of potential real-world applications. With wingspans as small as 2.5 centimeters \cite{Wood_2008}, FWMAVs have the potential of assisting in tasks such as artificial pollination, environmental monitoring, or search and rescue in small, difficult to reach areas. To accomplish such applications, fully autonomous flight of FWMAVs (which includes robust control and state estimation using an onboard electronics package) must be achieved. With the small scale of FWMAVs comes challenging dynamics that makes state estimation difficult. As vehicle size diminishes, the vehicle's dynamics scale accordingly. For example, rotational acceleration rate scales as $l^{-1}$ , resulting in the ability to perform rapid attitude changes, similar to saccades observed in flying insects and birds   \cite{Kumar_Michael_2012, 10.1093/icb/45.2.274, WissaAimy2022Btfs}. The vehicle dynamics are inherently unstable, requiring active control to perform corrective maneuvers which produces low frequency oscillation modes in FWMAV flights \cite{McLean_2003}. Furthermore, the induced oscillations from flapping wings generates additional disturbances to the observed sensor data. As a result, state estimators must carefully consider the vehicle's dynamics and characteristic modes to provide accurate estimations. While solutions have been derived for similar flight dynamics, an additional challenge for the design of any state estimation algorithm is accounting for the power, weight, and compute restrictions of these vehicles \cite{Aurecianus2018FlappingWM}. 

\begin{figure}[ht]
    \centering
    \includegraphics[width=0.85\linewidth]{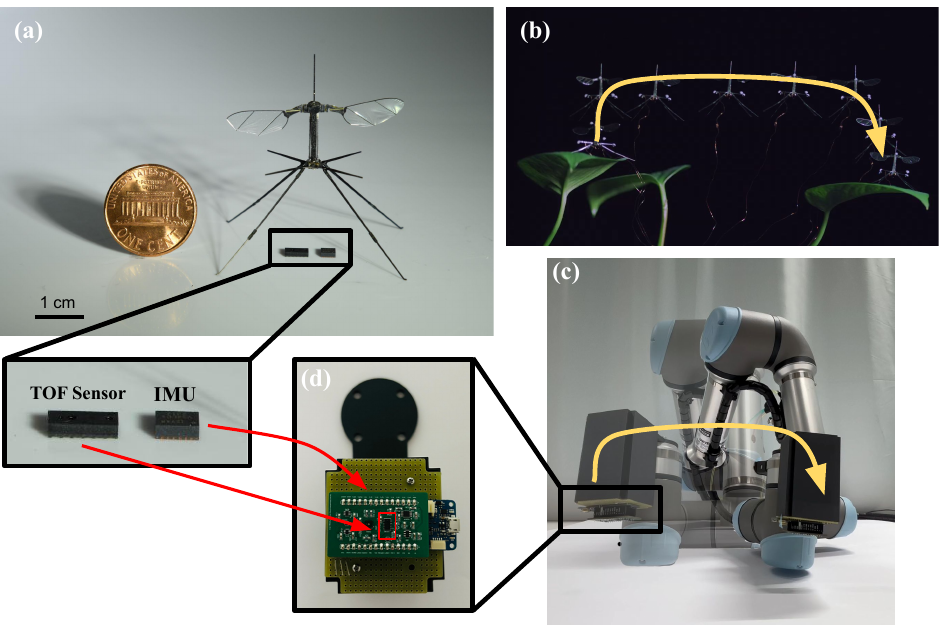}
    \caption{(a) The Harvard RoboBee, the target FWMAV, photographed with the IMU and ToF components utilized in the hardware experiments. (b) A RoboBee flight trajectory that the robot arm reproduces in (c), while carrying the IMU and ToF sensors to collect data. (d) The sensor package, mounted to the distal end of the arm, which contains the sensor components shown in (a) for state estimation. }
    \label{fig:motivation}
\end{figure}

The first demonstration of an insect-scale microrobot to carry its own weight solved the problem of flapping-wing propulsion at the micro-scale by driving the wings with piezoelectric bimorph actuators \cite{Wood_2008}. Due to the poor efficiency of motors at smaller scales (due to surface area to volume ratio scaling, resulting in a greater dominance of friction \cite{TRIMMER1989267}), alternate propulsion techniques for insect-scale FWMAV have been developed, including dielectric elastomer actuators and piezoelectric bending actuators \cite{Chen_Arase_Ren_Chirarattananon_2022, Ozaki_Ohta_Jimbo_Hamaguchi_2021}. This paper considers the Harvard RoboBee, an $81~\mathrm{mg}$ FWMAV shown in Figure \ref{fig:motivation} (a). The Harvard RoboBee leverages piezoelectric bimorph actuators to generate a thrust-to-weight ratio greater than unity, sufficient to power agile flight with some overhead available for a small payload ($40~\mathrm{mg}$). However, the vehicle's minimal payload, along with fast vehicle dynamics, has historically resulted in several tasks being performed off-board including computation and sensing. Specifically, current autonomous flights for the RoboBee rely on an array of external motion capture cameras  to provide the necessary state estimates required for feedback control \cite{McGill_Hyun_Wood_2022}. 

Research in control autonomy for larger systems has demonstrated onboard localization using a wide array of sensors, such as cameras and scanning laser range finders \cite{8444257} \cite{9811757}. However, the considered vehicles are one to two order magnitudes larger than the RoboBee, which makes several of the considered sensors infeasible as they do not satisfy the RoboBee's limited weight and power constraints. Nonetheless, biology offers evidence of a plethora of unique sensory mechanisms that enable birds and insects to maintain stable flight orientations \cite{doi:10.1126/science.1133598, doi:10.1126/scirobotics.abl6334, de_Croon_Dupeyroux_De_Wagter_Chatterjee_Olejnik_Ruffier_2022}. Thus, past research has similarly demonstrated the possibility of onboard sensor feedback for RoboBee control. Fuller et. al. considered several options for enabling stable orientation control for the RoboBee, using an onboard gyroscope and a bio-inspired ocelli (horizon detection sensor) \cite{Fuller_Helbling_Chirarattananon_Wood_2014, Fuller_Sands_Haggerty_Karpelson_Wood_2013}. Helbling et. al. demonstrated onboard integration of a Time-of-Flight sensor for altitude estimation for the RoboBee \cite{Helbling_Fuller_Wood_2017}. These prior research efforts informed the survey and selection of sensor components as shown in Figure \ref{fig:motivation} (a). However, the mechanical complexity of the RoboBee makes the integration and evaluation of potential sensing packages and algorithms challenging. As a result, the scale, manufacturing complexity, and limitations of the RoboBee prevented the prototyping of advanced estimation algorithms. This challenge along with computational limitations led to the use of relatively simple state estimation algorithms that relied on techniques like gyroscopic integration which produced significant drift up to 6 degrees in orientation estimation after only a 2 second flight experiment \cite{Fuller_Helbling_Chirarattananon_Wood_2014}.
\begin{figure*}[htbp]
    \centering
    \includegraphics[width=\textwidth]{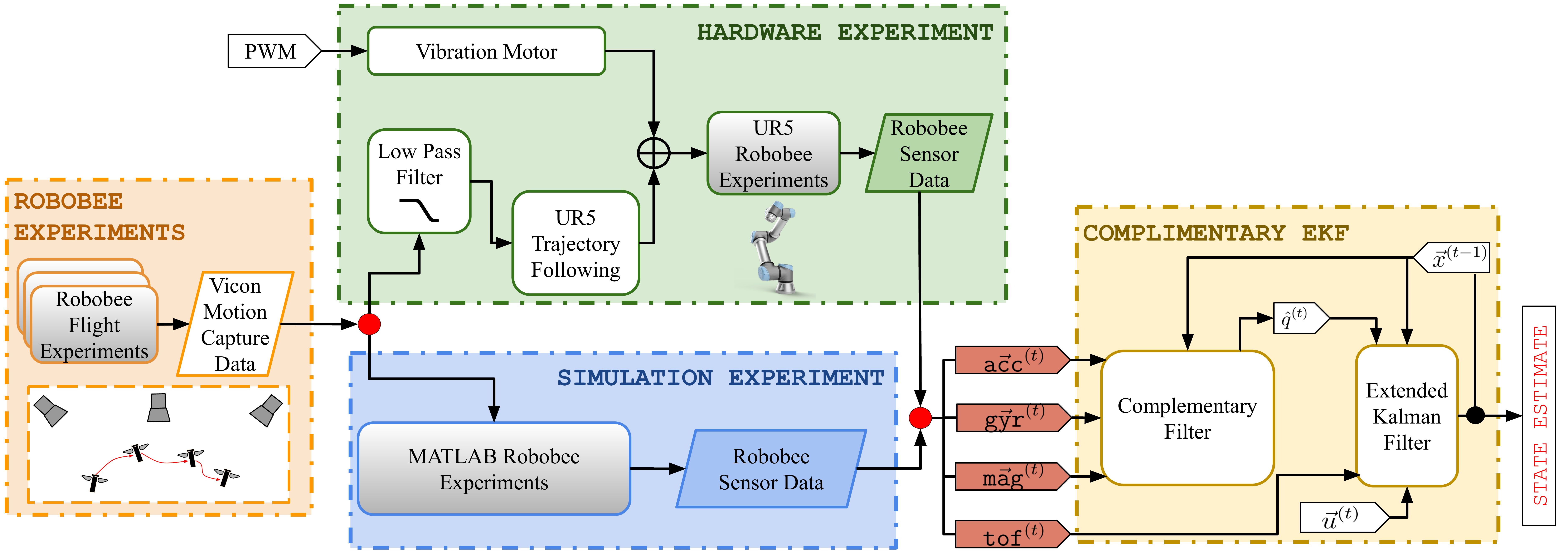}
    \caption{Block diagram illustrating the overall pipeline for the evaluation of state estimation algorithms. Based on RoboBee flight trajectory data, which is captured with a suite of Vicon motion capture cameras, a candidate onboard sensor package is evaluated by reproducing the same flight trajectories in hardware (or simulation). Subsequently, the sensor data is fed into the state estimation algorithm which utilizes a complementary filter and EKF to estimate the orientation and altitude of the vehicle.}
    \label{fig:pipeline}
\end{figure*}
These challenges motivated the development of a lightweight sensing suite, which integrates several of the components listed above, and a novel state estimation algorithm based on an Extended Kalman Filter by Talwekar et al. This algorithm simultaneously estimates pitch, altitude, and translational velocity \cite{9811918} for extended durations (20 seconds). To evaluate the proposed sensing package while avoiding the constraints imposed by insect-scale FWMAVs, Talwekar used an offboard experimental setup, in which he waved the sensing package by hand inside a motion capture arena to collect ground truth data for evaluation. Although this testing pipeline allowed for rapid evaluation and iteration, the dynamics of actual FWMAV flights are complex and introduce several sources of noise that were not reflected in their evaluation procedure. 

Although the majority of the previous sensing research for insect-scale FWMAVs do not explore more advanced state estimation algorithms, researchers have investigated novel estimators for vehicles with similar noisy dynamics. Mahony et. al derived a fundamental filtering architecture that applies to rotation matrices and similarly Madgwick et al. proposed a gradient-descent complementary filter for attitude estimation for quad-rotors \cite{Mahony_Hamel_Pflimlin_2008,Madgwick_Harrison_Vaidyanathan_2011}. In particular, researchers have explored algorithms for oscillating environments using filter architectures such as Cascaded Complementary Filters (CCF) and Complementary Kalman Filters \cite{Chiella_Teixeira_Pereira_2019, C_Jain_2016}. While these papers consider a similar challenge (i.e., state estimation for aerial robots), the performance of each algorithm is validated in a simulated environment, with minimal, if any, experiments on hardware. Given the notable hardware constraints, however, performing thorough hardware experiments to validate accuracy of potential state estimation algorithms and sensors under FWMAV flight dynamics is important for motivating the substantial efforts needed to bring the compute and sensors onboard. However, given the hardware complexity in the manufacturing of vehicles like Harvard's RoboBee, evaluate state estimation algorithms in hardware settings is quite challenging. Thus, given insect-scale FWMAV's limited accessibility and complex fabrication process, research has also been invested into developing methods of approximating RoboBee dynamics and controls for hardware-in-the-loop testing of more complicated software solutions that can only be solved with experimentation. Chen, et. al presented a method of simulating FWMAV dynamics on quadrotors to create a more reproducible experimental setup for software control/state estimation solutions \cite{Chen_Fuller_Dantu_2017}.
While this example successfully reproduces RoboBee dynamics by mapping the control inputs of a FWMAV to quadrotor inputs, they do not simulate the body oscillation dynamics that makes state estimation for FWMAVs particularly challenging. Effective evaluation of potential state estimation algorithms and sensors requires a more complete reproduction of FWMAV flight dynamics. Specifically, we seek to incorporate a richer suite of dynamics, including noise from various oscillation modes, in addition to reproducing gross trajectories in order to subject sensors and algorithms to more realistic flight scenarios. Consequently, in this paper, we not only present a minimal sensor suite and a simplified state estimation algorithm for FWMAVs, but also design a hardware-in-the-loop pipeline to characterize estimation performance. This pipeline recreates FWMAV flight experiments (Figure \ref{fig:motivation} (b)) using a UR5e (Figure \ref{fig:motivation} (c)) and adds realistic noise observed in flight experiments. This setup allows for rapid evaluation of algorithm accuracy on the considered hardware sensor components (Figure \ref{fig:motivation} (d)), achieving more accurate state estimation results than previous efforts.

\subsection{Our Contributions}
In this paper, we propose a comprehensive state estimation algorithm and sensor suite that satisfies the RoboBee's hardware constraints. Although we do not deploy the suite onboard the robot, we constrain hardware choices to those with opportunities for miniaturization. Specifically, we limit sensor selection to those that would fit within the approximately $50~\mathrm{mg} $ mass budget. The sensor suite incorporates an off-the-shelf nine-axis inertial measurement unit (IMU) with an accelerometer, magnetometer, and gyroscope (InvenSense ICM20948) and a time-of-flight (ToF) sensor (VL6180), shown in Figure~\ref{fig:motivation} (a), that provides sensor measurements for a Complementary Extended Kalman Filter (CEKF), based on RoboBee dynamics. Additionally, we develop a testing pipeline using both simulation and hardware experiments, leveraging RoboBee flight experiments (as outlined in Figure \ref{fig:pipeline}) to validate that the accuracy of the algorithm does not degrade with the presence of body oscillatory modes. The Supplemental Video provides an overview of replayed trajectories utilized in this work as well as the mechanical noise generator designed to induce body oscillatory dynamics on the sensor suite. We observe that under RoboBee flight dynamics, the CEKF achieves state estimates with less than $1^\circ$ RMSE error in orientation and less than $2~\mathrm{mm}$ error in altitude with 16-bit fixed-point precision.  Additionally, we approximate that a cycle of the proposed CEKF will require less than 11 microseconds on an onboard microcontroller for computations, which satisfies real-time latency requirements for the proposed state estimation algorithms.

The rest of the paper is organized as follows.
In Section~\ref{sec: CEKF} we derive the Complementary Extended Kalman Filter architecture for state estimation and  evaluate the performance on simulated sensor data generated from open-loop RoboBee flight experiments. In Section~\ref{sec: component selection} we select sensors that meet the SWaP constraints of the vehicle and characterize their performance. Section~\ref{sec: experiments} presents our hardware-in-the-loop evaluation approach and the performance of the sensor suite and algorithm with RoboBee closed-loop dynamics. Finally, given the real-time latency requirement of state feedback for controllers, section \ref{sec: discussion} evaluates the required floating point operations (FLOPS) and the rate at which CEKF accuracy degrades for less precise floating point bit representations.  

\section{Complementary Extended Kalman Filter} \label{sec: CEKF}
State estimation challenges for FWMAVs are difficult given the unique nature of the system-induced noise that is a combination of high frequency operation of the piezoelectric actuators and the wing-beat body oscillation dynamics described in Section~\ref{sec: intro}. In order to mitigate the effects of system-induced noise experienced during FWMAV flight experiments, a state (orientation and altitude) estimation algorithm that leverages both sensor measurements and a system model that is grounded in first principles is required. Thus, we propose a Complementary Extended Kalman Filter architecture that is built on a sensor suite composed of a nine-axis IMU and ToF sensor in order to estimate the orientation and altitude of the vehicle.


\subsection{Problem Formulation}
We formulate the state estimation problem as a standard state space model. We consider a sensor package that includes a nine-axis IMU (which includes an accelerometer, a gyroscope, and a magnetometer) and a ToF sensor that measures the proximity of the vehicle to a nearby surface (in this case, given the orientation of the ToF sensor in the body-attached coordinate frame, this is ground). Consequently, we can represent our measurement vector as follows at time $t$: $\vv{\rho}^{(t)} = \begin{bmatrix}
    \vv{\texttt{acc}}^{(t)} & \vv{\texttt{gyr}}^{(t)} & \vv{\texttt{mag}}^{(t)} & \texttt{tof}^{(t)} 
\end{bmatrix}^T$. Additionally, note that each $\vv{\texttt{acc}}^{(t)}, \vv{\texttt{gyr}}^{(t)}, \vv{\texttt{mag}}^{(t)} \in \mathbb{R}^3$ where, for example: $\vv{\texttt{acc}}^{(t)} = \begin{bmatrix}
    \texttt{acc}^{(t)}_x & \texttt{acc}^{(t)}_y & \texttt{acc}^{(t)}_z 
\end{bmatrix}$. 

Moreover, the RoboBee is controlled through commanded thrusts $F_T$ and torques $\Vec{\tau}^{(t)}$ in order to allow for full control authority of the heading and altitude of the vehicle \cite{McGill_Hyun_Wood_2022}. We define the control input mathematically as follows: $\vv{u}^{(t)} = \begin{bmatrix}
    \tau_x & \tau_y & \tau_z & F_T
\end{bmatrix}^T$, which is shown in Figure \ref{fig:RoboBee_diagram}.

Given the measurements, $\vv{\rho}^{(t)}$, and control inputs $\vv{u}^{(t)}$, we derive a state estimation algorithm that attempts to estimate both the orientation $\vv{q}^{(t)}$ and altitude $\zeta^{(t)}$ of the vehicle. Motivated by the fact that the majority of the RoboBee's flight experiments have minimal orientation variation, we define the heading of the vehicle as an $XYZ$ sequence of Euler angles $\vv{q}^{(t)} = \begin{bmatrix}
    \phi^{(t)} & \theta^{(t)} & \psi^{(t)}
\end{bmatrix}^T$ given that we will not encounter orientations with singularities \cite{Fuller_Helbling_Chirarattananon_Wood_2014}. Additionally, the complete rotation matrix between the world coordinate frame $\{W\}$ and the vehicle's body-attached coordinate frame $\{B\}$ can be recovered as follows: $R_{\{B\}}^{\{W\}} (\vv{q}^{(t)}) = R_{x}(\phi) R_y(\theta) R_z (\psi)$, where $R_{x}(\phi)$, $R_{y}(\phi)$, $R_{z}(\phi)$ are each rotation matrices about the axes of the rotating coordinate frame. Therefore the complete estimated state vector is defined as follows $\vv{s}^{(t)} = \begin{bmatrix}
    \vv{q}^{(t)} & \zeta^{(t)}
\end{bmatrix}^T$.

Finally, for conciseness, in the derivations below we use $s_{\theta}$, $c_{\theta}$, and $t_{\theta}$ to represent $\sin(\theta)$, $\cos(\theta)$, and $\tan(\theta)$, respectively.

\subsection{Cascaded Complementary Filter}
Complementary filters for IMU-based state estimations are one of the most widely used approaches in robotics \cite{inproceedings}. They offer a method of leveraging the complementary nature of nine-axis IMUs to provide state estimation that does not require any system modelling. Specifically, while gyroscopes are good at detecting changes in orientation over short periods, they suffer from drift/bias over time. Accelerometers and magnetometers, on the other hand, are much less susceptible to biases and drift that affect gyroscope measurements but are more affected by transient forces and magnetic disturbances which in turn makes detecting changes in orientation over short periods of time difficult.

In several proposed complementary filter architectures, the selection of various filter gain values and parameters are fragile \cite{Mahony_Hamel_Pflimlin_2008} \cite{Madgwick_Harrison_Vaidyanathan_2011}. Given the nature of FWMAV flights, tuning these parameters can be quite challenging. In this paper, we consider a novel Cascaded Complementary Filter proposed by Narkhede, et. al that employs both nonlinear and linear versions of the complementary filter within one framework \cite{Narkhede_Poddar_Walambe_Ghinea_Kotecha_2021}. The authors showed that the distinct advantage of the proposed architecture is that the performance of the estimator is not as sensitive and fragile to the selected parameters and allows for much easier tuning.

The structure of the CCF consists of two input signals $\vv{q}_1$ and $\vv{q}_2$, which are low and high frequency noise-corrupted versions of the true orientation signal $\vv{q}$. The low frequency signal, $\vv{q}_1$, can be directly computed from the gyroscope measurements:
\begin{align}
    \vv{q}_1^{(t)} &= \int_0^t \vv{\texttt{gyr}} ~ dt
\end{align}

For the high frequency signal estimate, $\vv{q}_2$, we leverage trigonometric relationships to derive an estimate for the system orientation using the accelerometer and magnetometer as follows:
\begin{align}
    \vv{q}_2^{(t)} &= \begin{bmatrix}
        \tan^{-1} \left( \frac{\texttt{acc}_y}{\texttt{acc}_z} \right) \\
        \tan ^{-1} \left( \frac{-\texttt{acc}_x}{\texttt{acc}_y s_{\phi} + \texttt{acc}_z c_{\phi}} \right) \\
        \tan ^{-1} \left( \frac{\texttt{mag}_z s_{\phi} - \texttt{mag}_y c_{\phi}}{\texttt{mag}_x c_{\theta} + \texttt{mag}_y s_{\theta} s_{\phi} + \texttt{mag}_z s_{\theta} c_{\phi}} \right)
    \end{bmatrix}
\end{align}

Finally, the linear complementary filter output is a weighted average based on a low pass and high pass filter:
\begin{align}
    \hat{q}^{(t)} &= \vv{q}_1^{(t)} \alpha  + \vv{q}_2^{(t)} (1 - \alpha)
    \label{eq: lin comp}
\end{align}

The reason CCF is well suited for noisy dynamics, such as those observed during FWMAV flight experiments, is because the CCF additionally uses a nonlinear complementary filter to reduce the steady-state error and compensate for the varying gyroscope drift. The resulting $\hat{q}$ is derived as follows, where $K_P$ and $K_I$ are the the proportional and integral gain, respectively:
\begin{align}
    \hat{q}^{(t)} &= \dfrac{1}{s} \left[\vv{\texttt{gyr}}^{(t)} + \left(K_P + \frac{K_I}{s}\right) \left(\vv{q}_2^{(t)} - \hat{q}^{(t)}\right) \right]
    \label{eq: nonlin comp}
\end{align}
Note that $K_P$, $K_I$, and $\alpha$ are constants chosen through trial and error, however as Narkhede claimed, we also observe that the performance of the CCF is robust to a wide range of gain values which is useful for FWMAV state estimation applications \cite{Narkhede_Poddar_Walambe_Ghinea_Kotecha_2021}. By substituting the result from equation \ref{eq: nonlin comp} (non-linear bias-correcting complementary filter output) for $\vv{q}_1^{(t)}$ in equation \ref{eq: lin comp} (linear complementary filter) we can mathematically express the CCF architecture in algebraic form as follows:

\begin{equation}
    \begin{split}
    \hat{q}^{(t)} = &\frac{\alpha s^2}{s^2 + \alpha K_P s + \alpha K_I} \left(\frac{\vv{\texttt{gyr}}^{(t)}}{s}\right) \\ &+ \frac{(1- \alpha) s^2  + \alpha K_P s + \alpha K_I}{s^2 + \alpha K_P s + \alpha K_I}  \vv{q}_2^{(t)}
    \end{split}
\end{equation}

The above expression allows for accurate estimation of the system's orientation given that the low-pass filtering of the accelerometer measurements and high-pass filtering of the gyroscope measurements together help to estimate the attitude angles across all frequencies of interest.

\subsection{Extended Kalman Filter} \label{subsec: ekf}

\begin{figure}
    \centering    \includegraphics[width=\columnwidth]{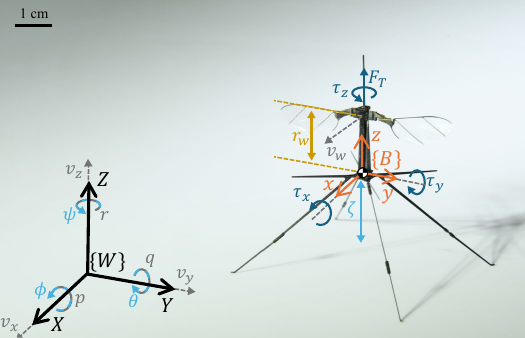}
    \caption{The Harvard RoboBee; the state described via the roll ($\phi$), pitch ($\theta$), yaw ($\psi$), and altitude ($\zeta$) in the world coordinate frame $\{W\}$, and the control inputs $\tau_x$, $\tau_y$, $\tau_z$, and $F_T$ are described in the body coordinate frame $\{B\}$.}
    \label{fig:RoboBee_diagram}
\end{figure}
As motivated earlier, the purpose of the EKF is to utilize the existing understanding of RoboBee dynamics to mitigate the potential drift in the state estimation. This understanding can be derived from the extensive body of work analyzing the aerodynamics of insect flight, which motivated the development FWMAV systems such as the Harvard RoboBee \cite{sane_dickinson_2002}. When Kalman derived the original EKF, although global convergence to the true state is not guaranteed, local convergence is guaranteed given that the system is observable \cite{10.1115/1.3662552}. As a result, we design the RoboBee's EKF such that we can ensure observability over the orientation and altitude states to guarantee accurate estimations.

The output of the EKF is the orientation and altitude of the vehicle, thus we define $\vv{\xi}^{(t)} = \begin{bmatrix}
    \vv{q}^{(t)} & \zeta^{(t)}
\end{bmatrix}$. The orientation and altitude dynamics of the RoboBee are predominantly influenced by two main factors: (1) the commanded control thrust and torques, and (2) the accompanying drag force and torque produced by the wings. These dynamics are also fundamentally governed by the RoboBee's inertial properties, including its mass and moments of inertia.

The control inputs for the RoboBee has been thoroughly studied through the iterations of the RoboBee physical design. Although the control space of the actuation mechanism for the Harvard RoboBee is high dimensional, McGill et. al. developed a quasi-static non-linear mapping between the commanded voltages and the resulting torques ($\Vec{\tau}_U^{\{B\}} = \begin{bmatrix}
    \tau_x & \tau_y & \tau_z
\end{bmatrix}$) and thrust ($\Vec{F}_T^{\{B\}} = \begin{bmatrix}
    0 & 0 & F_T
\end{bmatrix}$) that allows for full control over the orientation of the RoboBee \cite{McGill_Hyun_Wood_2022}. Representing FWMAV flight dynamics using torques and thrust is much more attractive than handling the high dimensional control space available to vehicles like the RoboBee. Thus, we choose to represent our control input as follows $\vv{u}^{(t)} = \begin{bmatrix}
 \vv{\tau}_U^{\{B\}} & F_T^{\{B\}}   
\end{bmatrix}$. As annotated in Figure \ref{fig:RoboBee_diagram}, the induced torques and thrust are represented in the body coordinate frame.

A side-effect of utilizing wings to generate thrust and torques is that translational accelerations induce drag on the vehicle that affects the dynamics. Previous work has shown that the force produced from drag on flapping wings is approximately linearly related to the lateral velocity of the wings $v_w$ (Figure \ref{fig:RoboBee_diagram}) by a factor of $b_w$ (drag constant, as shown in Table \ref{tab: parameters}). The lateral velocity of the wings, $v_w$, is a combination of two components. The first component is the linear velocity along the $x$-axis of the body coordinate frame $v_x^{\{B\}}$, which can be computed by projecting the velocity of the vehicle in the world coordinate frame $\vv{v}^{\{W\}}$ into the body coordinate frame as follows. 
\begin{align}
v_x^{\{B\}} = \begin{bmatrix}
        c_{\psi} c_{\theta} \\ c_{\theta}s_{\psi} \\
        s_{\theta}
\end{bmatrix}^T \vv{v}^{\{W\}}    
\end{align}
The second component is the velocity produced at the midpoint of the wings by the angular velocity around the body frame's $y$-axis. This value can be computed by projecting the world coordinate frame angular velocity, $\vv{\omega}^{\{W\}}$, onto the body coordinate frame $y$-axis and multiplying by the distance between the midpoint of the wings and the center of mass (CM), $r_w$, as follows:
\begin{align}
    v_2^{\{B\}}= \begin{bmatrix}    c_{\psi}s_{\theta}s_{\phi} - s_{\psi}c_{\phi} \\  s_{\psi}s_{\theta}s_{\phi} - c_{\psi}c_{\phi} \\ c_{\theta}c_{\phi}
\end{bmatrix}^T r_w \vv{\omega}^{\{W\}}
\end{align}

Thus, given that we define $v_w = v_x^{\{B\}} + v_2^{\{B\}}$, we can directly compute the produced drag force $\vv{F}^{\{B\}}_D$ and induced torque $\vv{\tau}^{\{B\}}_D$ as follows:
\begin{align}
    \vv{F}^{\{B\}}_D = \begin{bmatrix}
    -b_w v_w & 0 & 0
    \end{bmatrix}^T \\
    \vv{\tau}^{\{B\}}_D = \begin{bmatrix}
    0 &  r_w (b_w v_w) & 0
\end{bmatrix}
\end{align}

We derive the total force and torque vectors acting on the RoboBee during flight in the body coordinate frame as the sum of the control input and drag: $\vv{F}^{\{B\}} = \vv{F}_D^{\{B\}} + \vv{F}_T^{\{B\}}$ and $\vv{\tau}^{\{B\}} = \vv{\tau}_D^{\{B\}} + \vv{\tau}_U^{\{B\}}$.

Given that the estimation of the output states $\vv{q}$ and $\zeta$ require the forces and torques to be modelled, the state space considered in the EKF must be a ten-dimensional vector that includes the Euler angles ($\vv{q} = \begin{bmatrix}
    \phi & \theta & \psi
\end{bmatrix}$), the angular velocities in the world frame ($\vv{\omega}^{\{W\}} = \begin{bmatrix}
    p & q & r
\end{bmatrix}$),  the linear velocities in the world frame ($\vv{v}^{\{W\}} = \begin{bmatrix}
    v_x & v_y & v_z
\end{bmatrix}$), and the altitude $\zeta$. Figure \ref{fig:RoboBee_diagram} is a diagram of the physical attributes and dynamics for Harvard's RoboBee. Given $\vv{s}^{(t)}$ and $\vv{u}^{(t)}$ we derive our nonlinear discrete system dynamic model as follows:
\begin{align}
    \vv{s}^{(t+1)} &= f(\vv{s}^{(t)}, \vv{u}^{(t)}) \\
    &=\vv{s}^{(t)} + \Delta\begin{bmatrix}
        \vv{\omega}^{\{W\}} &
        \dfrac{1}{I}\vv{\tau}^{\{W\}} &
        \dfrac{1}{m}\vv{F}^{\{W\}} &
        v^{\{W\}}_z
    \end{bmatrix}^T 
\end{align}
Note that $I$ is the moment of inertia about each axis and $m$ is the mass of the vehicle, as listed in Table \ref{tab: parameters}. Additionally, given that we assume that RoboBee operates near hovering set points, we approximate the derivative of the Euler angles to the angular velocity vector to reduce excess computations. Furthermore note that the torques and forces in $f(\vv{s}^{(t)}, \vv{u}^{(t)})$ are the body coordinate torques and forces transformed by $R^{\{W\}}_{\{B\}}$.

Additionally we define $H$, such that $\vv{\xi}^{(t)} = H \vv{s}^{(t)}$. We compute the time variant observability matrix as follows: $\mathcal{O} = \begin{bmatrix}
    H & H \mathcal{J} & H \mathcal{J}^2 & \cdots & H \mathcal{J}^9
\end{bmatrix}^T$, where $\mathcal{J}$ is the jacobian of the state transition function ($f(\vv{s}^{(t)}, \vv{u}^{(t)})$). Although $\mathcal{O}$ is not full rank for all time $t$, the columns corresponding to the output state space are linearly independent proving partial observability and guaranteeing local convergence to the true state for the orientation $\vv{q}^{(t)}$ and $\zeta^{(t)}$ based on the measurement vector $\vv{z}^{(t)} = \begin{bmatrix}
    \hat{q}^{(t)} & \hat{\zeta}^{(t)}
\end{bmatrix}$, which is derived from the output of the complementary filter, $\hat{q}^{(t)}$, and the ToF measurement, $\texttt{tof}^{(t)}$. We derive the measurement estimate of the vehicles altitude $\hat{\zeta}^{(t)}$ as follows:
\begin{align*}
    \hat{\zeta}^{(t)} &= \texttt{tof}^{(t)} c_{\hat{\phi}^{(t)}} c_{\hat{\theta}^{(t)}}
\end{align*}
Given the above model for the RoboBee, implementing an Extended Kalman Filter follows directly from \cite{Fuller_Helbling_Chirarattananon_Wood_2014}. Given $f(\vv{s}^{(t)}, \vv{u}^{(t)})$, we first predict the next state $\hat{x}^{(t+1)}$. Additionally, we linearize the dynamics around $x^{(t)}$ by computing $\mathcal{J}$ in order to find the updated state uncertainty $P^{(t)}$ as follows:
\begin{align*}
    \hat{P}^{(t+1)} &= \mathcal{J} P^{(t)} \mathcal{J}^T + Q
\end{align*}
where $Q$ is our model uncertainty and $R$ is our measurement uncertainty. 

Subsequently, we compute the Kalman gain for the correction step as follows:
\begin{align}
    K^{(t+1)} &= \hat{P}^{(t+1)} H^T (H\hat{P}^{(t+1)}H^T + R)^{-1}
    \label{eq: kalman gain}
\end{align}
Finally, given the updated Kalman gain we can compute our estimated state and estimated uncertainty given our measurement estimate $\vv{z}^{(t+1)}$:
\begin{align}
    \vv{s}^{(t+1)} &= \hat{x}^{(t+1)} + K(\vv{z}^{(t+1)} - H \hat{x}^{(t+1)}) \\
    P^{(t+1)} &= \hat{P}^{(t+1)} - KH\hat{P}^{(t+1)}
\end{align}
\subsubsection{Parameter Tuning}
In any state estimation application the challenge of selecting parameters to yield performance that generalizes to as many trajectories and dynamics is fundamental. The complementary filter parameters ($K_P$, $K_I$, and $\alpha$) were selected by minimizing the estimation error on 20 reference open loop RoboBee flight experiments. Subsequently, we define the measurement uncertainty matrix $R$ in RoboBee's EKF based on the datasheet noise characteristics of the sensors as listed in Section \ref{sec: component selection}. Finally we tune the model uncertainty matrix, $Q$, through trial and error in simulation. For the results listed below we utilized $Q = \diag{\begin{bmatrix}
    0.1 & 0.1 & 0.1 & 1 & 1 & 1 & 0.0025 & 1 & 1 & 1
\end{bmatrix}}$ and $R = \diag{\begin{bmatrix}
    0.07& 0.07& 0.07& 0.002
\end{bmatrix}}$. These uncertainty parameter values reflect that we are slightly more confident in our sensor state estimates, however the RoboBee's model helps ensure that the predicted state is in agreement with the sensor observations minimizing potential drift.

\subsection{Simulation Evaluation}
In addition to introducing a RoboBee-compatible state estimation algorithm, we propose alternative methods to evaluate such implementations prior to bringing them onboard the RoboBee. The first evaluation technique involves generating simulated measurement vector $\vv{\rho}$ based on real RoboBee flight experiment trajectory data in MATLAB. The simulated sensor measurements for the nine-axis IMU and the ToF were generated based on 80 open-loop flight trajectories on the RoboBee. The chosen physical parameters coincide with the true values for RoboBee hardware and are summarized in Table \ref{tab: parameters}. 

\begin{table}[hbt]
    \centering
    \caption{RoboBee Parameter Values}
    \begin{tabular}{|c|l|c|c|}
        \hline 
         Symbol          & \makecell{Meaning}  & Value & Unit\\ 
         \hline
         \hline
         $m$             & \makecell{Mass} & $8.6\times10^{-2}$ & g \\ \hline
         $I_{xx}$  & \makecell{Moment of Inertia \\ $x$-axis} & $1.42\times10^{-6}$ & $\text{g}\times\text{m}^2$ \\ \hline
         $I_{yy}$  & \makecell{Moment of Inertia \\ $y$-axis} & $1.34\times10^{-6}$ & $\text{g}\times\text{m}^2$ \\ \hline
         $I_{zz}$  & \makecell{Moment of Inertia \\ $z$-axis} & $4.5\times10^{-7}$ & $\text{g}\times\text{m}^2$ \\ \hline     
         $b_w$             & \makecell{Drag coefficient} & $2\times10^{-4}$ & $\text{m}^{-1}$           \\ \hline
         $r_w$             & \makecell{Distance between \\ wings and COM}& $9$ & mm           \\ \hline
    \end{tabular}
    \label{tab: parameters}
\end{table}

We evaluate the performance of the CEKF by evaluating the Root Mean Square Error (RMSE) between the estimated state and the ground truth reference trajectory as follows:
\begin{align*}
    RMSE &= \sqrt{\dfrac{1}{T} \sum_{t=1}^T (x^{(t)} - \Bar{x}^{(t)})^2}.
\end{align*}

The CEKF RMSE for the 80 open-loop flights is shown in Table \ref{tab: sim_results}. We observe approximately 6 degree estimation error for each Euler angle with respect to the original motion capture data. While this is higher than state estimation accuracy requirements, it is important to note that open-loop flight experiments on the RoboBee are quite unpredictable and produce significant orientation variation with an average fluctuation range of $\pm 36$ degrees about each axis. Consequently, we can conclude that the $\approx 6$ degree RMSE supports the conclusion that the CEKF framework is able to produce valid estimations in challenging dynamics, given the somewhat random trajectories observed in open-loop RoboBee flight experiments. Additionally, we observe that the altitude ($\zeta$) estimations are accurate with less than 2.5 mm RMSE, even in open-loop flight experiments.   

\begin{table}[hbt]
    \centering
    \caption{CEKF Performance RMSE in Simulation}
    \begin{tabular}{|c |c |c | c| c|}
    \hline
           &  Roll & Pitch & Yaw & Height \\ \hline \hline
         Mean RMSE &  5.27 $^{\circ}$ & 5.17 $^{\circ}$   & 6.01 $^{\circ}$   & 2.4 mm\\\hline
          Median RMSE &  4.23 $^{\circ}$  &  4.56 $^{\circ}$  &  6.54 $^{\circ}$  &  1.2 mm\\\hline
         Std Dev RMSE &  2.38 $^{\circ}$  &  3.71 $^{\circ}$ &   2.37 $^{\circ}$  &  0.6 mm\\\hline
    \end{tabular}
    \label{tab: sim_results}
\end{table}
\section{Component Selection}\label{sec: component selection}
In order to implement the aforementioned CEKF onto the RoboBee, we must select scale-appropriate sensors to provide the necessary state estimates. Given the limited payload and power capacity of insect-scale FWMAVs, it is important these peripheral components fit within strict mass and power constraints. In this paper, we select the Invensense ICM-20948 nine-axis IMU and the STMicroelectronics VL6180 ToF sensor to track attitude and altitude, respectively. Relevant parameters and specifications on these sensors are listed in Table \ref{tab: imu_specs} and Table \ref{tab: prox_specs}. Both sensors are able to provide reliable raw data readout in RoboBee-compatible packages. Variations of the Harvard RoboBee design have demonstrated a payload capacity up to 370 mg, and the total mass of the sensor components considered in this paper is approximately 45 milligrams, satisfying the weight constraint \cite{7353575, Jafferis_Helbling_Karpelson_Wood_2019}. Additionally, the package sizes of these sensor components are shown in Figure \ref{fig:motivation} (a).

\begin{table}[hbt]
    \centering
    \caption{ICM-20948 9 Axis IMU Relevant Parameters}
    \begin{tabular}{|l |r |}
    \hline
        Parameter &  Value \\\hline\hline
        Mass & 23 mg \\\hline
        Volume &  $3 \times 3 \times 1  ~\mathrm{mm^3}$\\\hline
        Power & $1.8\mathrm{V}, 5.6~\mathrm{mW}$ \\\hline\hline
        \multicolumn{2}{|l|}{\textbf{Sampling Rates}} \\\hline
        Gyroscope & 9 $\mathrm{kHz}$ \\\hline
        Accelerometer & 4.5 $\mathrm{kHz}$ \\\hline
        Magnetometer & 100 $\mathrm{Hz}$ \\\hline\hline
        \multicolumn{2}{|l|}{\textbf{Data Range}} \\\hline
        Gyroscope & $\pm$2000 $\mathrm{dps}$ \\\hline
        Accelerometer & $\pm$16 $\mathrm{g}$ \\\hline
        Magnetometer & $\pm$4900 $\mu$$\mathrm{T}$ \\\hline\hline
        \multicolumn{2}{|l|}{\textbf{Data Sensitivity}} \\\hline
        Gyroscope & 16.4 LSB / $\mathrm{dps}$ \\\hline
        Accelerometer & 2048 LSB / $\mathrm{g}$ \\\hline
        Magnetometer & 0.15 $\mu$$\mathrm{T}$ / LSB\\\hline
    \end{tabular}
    \label{tab: imu_specs}
\end{table}

\begin{table}[hbt]
    \centering
    \caption{VL6180 Proximity Sensor Relevant Parameters}
    \begin{tabular}{|l |r |}
    \hline
        Parameter &  Value \\\hline\hline
        Mass & 22 mg \\\hline
        Volume &  $4.8 \times 2.8 \times 1  ~\mathrm{mm^3}$\\\hline
        Power (at 50 Hz, 20 cm) & $2.8~\mathrm{V}, 21~\mathrm{mW}$ \\\hline
        Data Range & 20 $\mathrm{cm}$ \\\hline
        Sampling Rate (at 20 $\mathrm{cm}$) & 50 $\mathrm{Hz}$ \\\hline
        Data Sensitivity (at 20 $\mathrm{cm}$) & 0.78 $\mathrm{mm}$ / LSB  \\\hline
    \end{tabular}
    \label{tab: prox_specs}
\end{table}

\subsection{Nine-axis Inertial Measurement Unit}
The selected nine-axis IMU combines a three-axis gyroscope, three-axis accelerometer, and three-axis magnetometer into a $3~\mathrm{mm}~\times~3~\mathrm{mm}~\times~1~\mathrm{mm}$ package that weighs $23~\mathrm{mg}$ and consumes approximately 5.6mW with all sensors running continuously. Previous work has shown that sensors with these dimensions are compatible with insect-scale FWMAVs \cite{Fuller_Helbling_Chirarattananon_Wood_2014, 9811918}. However, the selected sensor incorporates programmable precision and range, measuring up to $\pm16~\mathrm{g}$  linear acceleration and $\pm2000~\mathrm{dps}$ angular velocity. Additionally, the on-chip magnetometer provides compass data ranging up to $4900~\mathrm{\upmu T}$, which bounds the strength of Earth's magnetic field ($\approx35~\mathrm{\upmu T}$), enabling orientation estimates. In 80 open-loop RoboBee flight experiments, the maximum observed translational acceleration was $14.13~\mathrm{m/s^2}$ and the maximum angular velocity was $1194.4~\mathrm{dps}$, thus the flight envelope for FWMAVs is within the measurement ranges of the selected sensors.

Furthermore, the ICM-20948 hosts an integrated Digital Motion Processor (DMP) that provides real time filtering, calibration, and processing of the raw sensor signals. In this work, we leverage the DMP which performs the role of the complementary filter in Figure \ref{fig:pipeline} which offers a maximum output rate of 225 Hz. 

\subsection{Proximity Sensor}
In addition to approximating orientation, the proposed sensing suite must include an estimate of relative position to a nearby surface (i.e., ground) to estimate altitude. The IR tranceiver-based ToF sensor is contained in a package size of $4.8~\mathrm{mm} \times 2.8~\mathrm{mm} \times 1$ weighing $22~\mathrm{mg}$ and consuming $21$mW power running at 50Hz. By orienting the device to face the ground, we obtain approximate altitude measurements which we adjusted to accommodate variations in orientation as outlined in Section \ref{subsec: ekf}. Combining an IR emitter and range sensor, the VL6180 has a sensor range of up to $20~\mathrm{cm}$ with millimeter precision and consumes $21~\mathrm{mW}$ of power during operation. While this range enables accurate altitude estimation within typical RoboBee flight trajectories, the sensor also has extended range capabilities with reduced precision and increased power consumption, allowing our implementation to be extended to broader flight envelopes.

Optical ToF sensors are often viewed as power intensive components due to the periodic laser pulses required to generate measurements. However, the VL6180 is designed to mitigate excess power usage by limiting current consumption through adjustable precision, range, SNR, and dynamic sampling rates. Additionally, the VL6180 leverages ToF-style sensing to reduce the impact of target reflectivity, further boosting its overall versatility.

As an alternative, a MEMS-based ultrasonic proximity sensor, InvenSense CH101, was considered due to its comparable size and ultra-low power operation. While these devices consume much less power than their optical counterparts, ultrasonic sensors are limited by poor sound output due to large acoustic impedance at the interface between the sensor and the load (air in this case). This can be mitigated by attaching an additional component as an acoustic interface, such as a tube or horn, to match the impedance. This improves sound energy transfer but introduces considerable weight and complexity to the design and integration.

\subsection{Hardware Package and System Integration}
\begin{figure}[ht]
    \centering
    \includegraphics[width=0.8\columnwidth]{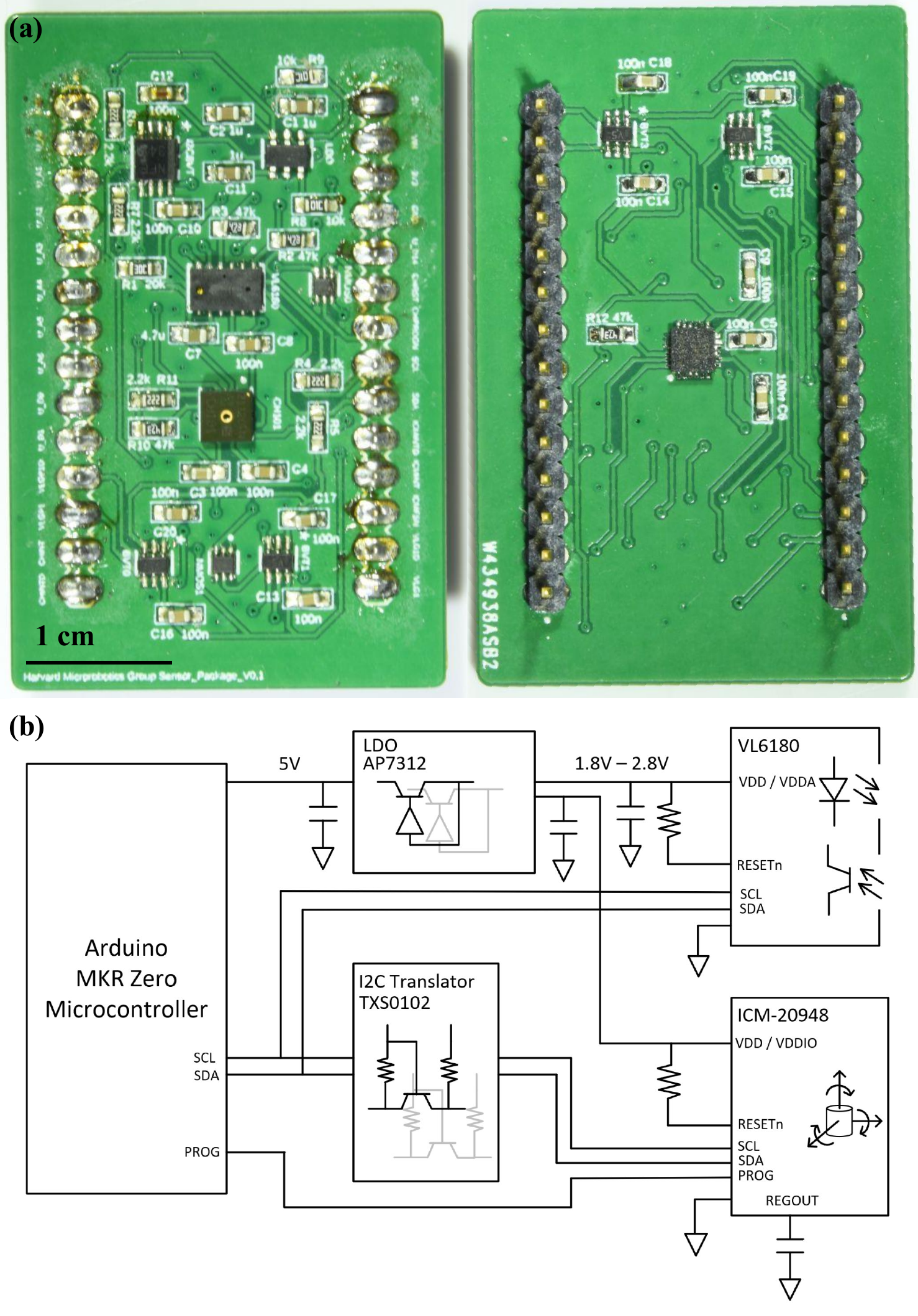}
    \caption{Prototype PCBs and schematic  for interfacing with the Invensense ICM-20948 nine-axis IMU and the VL6180 ToF sensor for hardware-in-the-loop testing.}
    \label{fig:pcbs}
\end{figure}

To facilitate rapid prototyping and testing, we fabricated rigid PCBs to mount the selected sensors and required peripherals as shown in Figure \ref{fig:pcbs}. To request and read out measurements, both sensors used a 400kHz I2C protocol with sampling rates set to 225Hz, which was the fastest output rate for the onboard DMP. While sampling at this rate is slower than the 500Hz used in the Vicon Motion Capture System, we show that the estimated position still remains within acceptable limits.

We also consider potential avenues to scale down the system to meet the mass requirements of FWMAVs. Several components included in the rigid PCBs can be omitted as only the power supply and I2C pins are required to communicate. The remaining required components include two decoupling capacitors, one close to each sensor to ensure workable supply voltages are available at all times. One other filter capacitor is required on the ICM-20948 to stabilize operation. The two sensors share the I2C communication channels, reducing the total wire count to four. The schematic in Figure \ref{fig:pcbs}(b) shows the minimum components required to implement our prototype and obtain state estimates. This can be decreased further by replacing the Arduino component with a smaller, low voltage microprocessor that can be mounted on the RoboBee. This change would eliminate the I2C translator component required to communicate between the 3.3V - 5V microcontroller board and the 1.8V - 2.8V sensors. 

\section{Experimental Evaluation} \label{sec: experiments}
We propose a hardware-in-the-loop testing pipeline (Figure \ref{fig:pipeline}) to 
enable direct evaluation of our state estimation algorithm under emulated RoboBee flight conditions. While redesigning RoboBee to integrate onboard sensors is an important step, it is a tangential problem to the actual state estimation algorithm. In addition, any redesign to accomodate additional sensing hardware must be informed by the design of the sensor package itself (i.e., performance, size, mass, ideal sensor placement). Consequently, in order to rapidly design and evaluate our proposed CEKF, we utilize a UR5 robot arm to reproduce previous closed-loop RoboBee flight experiments. The proposed sensor package is attached to the distal end of the arm, and thus we can directly compare the estimated state with ground truth from the arm kinematics.

\begin{figure}[ht]
    \centering
    \includegraphics[width=\columnwidth]{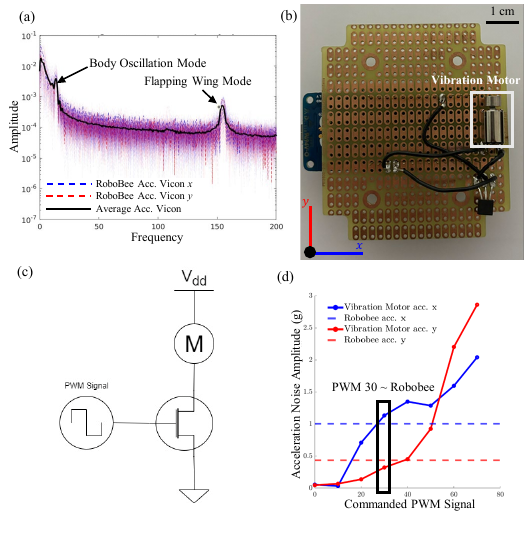}
    \caption{Characterization of RoboBee flight dynamics and device used to replicate induced noise in our hardware-in-the-loop experiments. (a) The frequency spectrum of the translational acceleration along the $x$ and $y$ axes for 80 open-loop RoboBee flight experiments, from external motion capture (Vicon) data. The frequency spectrum displays two primary modes of interest that contribute to oscillations: the ``body'' mode and ``flapping wing'' mode. (b) The vibration unit is designed to reproduce the body oscillation dynamics observed in (a) using a low frequency vibration motor. (c) A PWM-based control circuit is used to adjust the speed of the vibration motor to allow us to approximate the magnitude of the  oscillations shown in (a). (d) The amplitude of oscillations as a function of PWM duty cycle  relative to the amplitude of the body oscillation mode. }
    \label{fig:hardware_testing_characterization}
\end{figure}

As shown in Figure \ref{fig:hardware_testing_characterization} (a), the RoboBee experiences two prominent vibration modes: a body oscillation mode at approximately 13 Hz and a wingbeat mode at approximately 150 Hz. In this paper, the focus is on the body oscillation mode, given that the wingbeat mode can be mitigated with a  low pass filtered and the body oscillation mode is inherent to the closed-loop system dynamics (i.e., a consequence of both the body design and controller design). Trajectories from RoboBee flight experiments are fed into the UR5 built-in trajectory follower that is able to consistently produce the same trajectory with an average error of less than 1 degree for all three axes. Additionally, we superimpose the body oscillation mode through an attached vibration motor to the sensor board as shown in Figure \ref{fig:hardware_testing_characterization} (b). The vibration motor was selected to produce oscillations of varying magnitudes at 15 Hz. As shown in Figure \ref{fig:hardware_testing_characterization} (c), we utilize an NMOS transistor to control the speed of the vibration motor via PWM signals in order to allow for calibration of the produced noise to the body oscillation dynamics observed during RoboBee flight experiments. We calibrate the vibration unit based on the produced peak-to-peak amplitude of the oscillation in the acceleration along the $x$ and $y$ axes. From the reference RoboBee hovering experiment, we observe that the body oscillation mode induces approximately $1~\mathrm{g}$ acceleration along the $x$-axis and $0.5~\mathrm{g}$ acceleration along the $y$-axis. As seen in Figure \ref{fig:RoboBee_diagram}, the body frame $x$-axis is aligned with the wing velocity vector $v_w$ which supports the greater observed oscillation along the body frame $x$-axis compared to the $y$-axis. Additionally, there is significant coupling between different oscillation modes. We hypothesize that inducing body oscillations translationally along $x$ and $y$ will yield similar oscillation dynamics along other axes such as altitude and attitude. As shown in Figure \ref{fig:hardware_testing_characterization} (d), the PWM command that produces approximately the same dynamics of the body oscillation mode is 30\% duty cycle. Additionally, we can observe that once the vibration motor exceeds 50\% duty cycle the motor is much less stable and yields irregular noise behaviors compared to the lower duty cycles.

\subsection{Results}
\begin{table*}[t!]
\setlength{\extrarowheight}{3pt}
\centering
\caption{ Experimental CEKF Performance RMSE at different PWM duty cycles. The PWM signal corresponds to magnitude of noise induced upon sensor suite during experiments.}
\begin{tabular}{|c|c| c c c c c c c c |}
\hline
\multicolumn{1}{|c|}{} & \multicolumn{9}{c|}{CEKF Estimation RMSE}                                             \\ \hline 
& & PWM 0 & PWM 10 & PWM 20 & \textbf{PWM 30} & PWM 40 & PWM 50 & PWM 60 & PWM 70 \\ \hline \hline
\parbox[t]{2mm}{\multirow{4}{*}{\rotatebox[origin=c]{90}{\textit{\textbf{Hovering}}}}}   & Roll (deg) & 0.64 & 0.47& 0.52 & \textbf{0.64} & 0.88 & 1.29& 2.39 & 1.65 \\  
                             & Pitch (deg) & 0.16 & 0.17& 0.18 & \textbf{0.42} & 0.57 & 0.49& 2.59 & 4.16 \\ 
                             & Yaw (deg) & 1.64 & 1.23& 1.45 & \textbf{0.91} & 1.5 & 1.29& 1.88 & 1.94 \\  
                             & Altitude (mm) & 1.7 & 1.6 & 1.5 & \textbf{1.7} & 1.8 & 1.7 & 1.8 & 1.7 \\ \hline \hline
\parbox[t]{2mm}{\multirow{4}{*}{\rotatebox[origin=c]{90}{\textit{\textbf{Leaf Hopping}}}}}   & Roll (deg) & 0.68 & 0.97 & 0.90 & \textbf{0.76} & 0.99 &  2.66& 2.64& 2.20\\  
                             & Pitch (deg) & 0.76 & 0.47 & 0.43 & \textbf{0.47}& 1.63 & 1.29 & 1.46& 1.63\\ 
                             & Yaw (deg) & 2.83 & 2.89 & 2.59 & \textbf{2.61}& 2.54 & 2.98& 7.88& 7.39\\  
                             & Altitude (mm) & 1.9 & 2.1& 2.2 & \textbf{2.1} & 2.2 & 2.1& 1.9 & 2.1 \\ \hline                              
 \end{tabular}
 \label{tab:results}
\end{table*}    
Here, we consider two distinct RoboBee closed-loop flight trajectories: (1) a hovering trajectory, where the RoboBee takes off holds an orientation and altitude for approximately five seconds and lands, and (2) a leaf hopping trajectory, where the RoboBee takes off from a leaf, flies to another leaf, and lands.
As outlined in Table \ref{tab:results}, the CEKF is able to achieve RMSE errors of less than $3^\circ$ and $2.5~\mathrm{mm}$ for both reference trajectories with body oscillation dynamics similar to that of the Harvard RoboBee. 

\begin{figure}[ht]
    \centering
\includegraphics[width=\columnwidth]{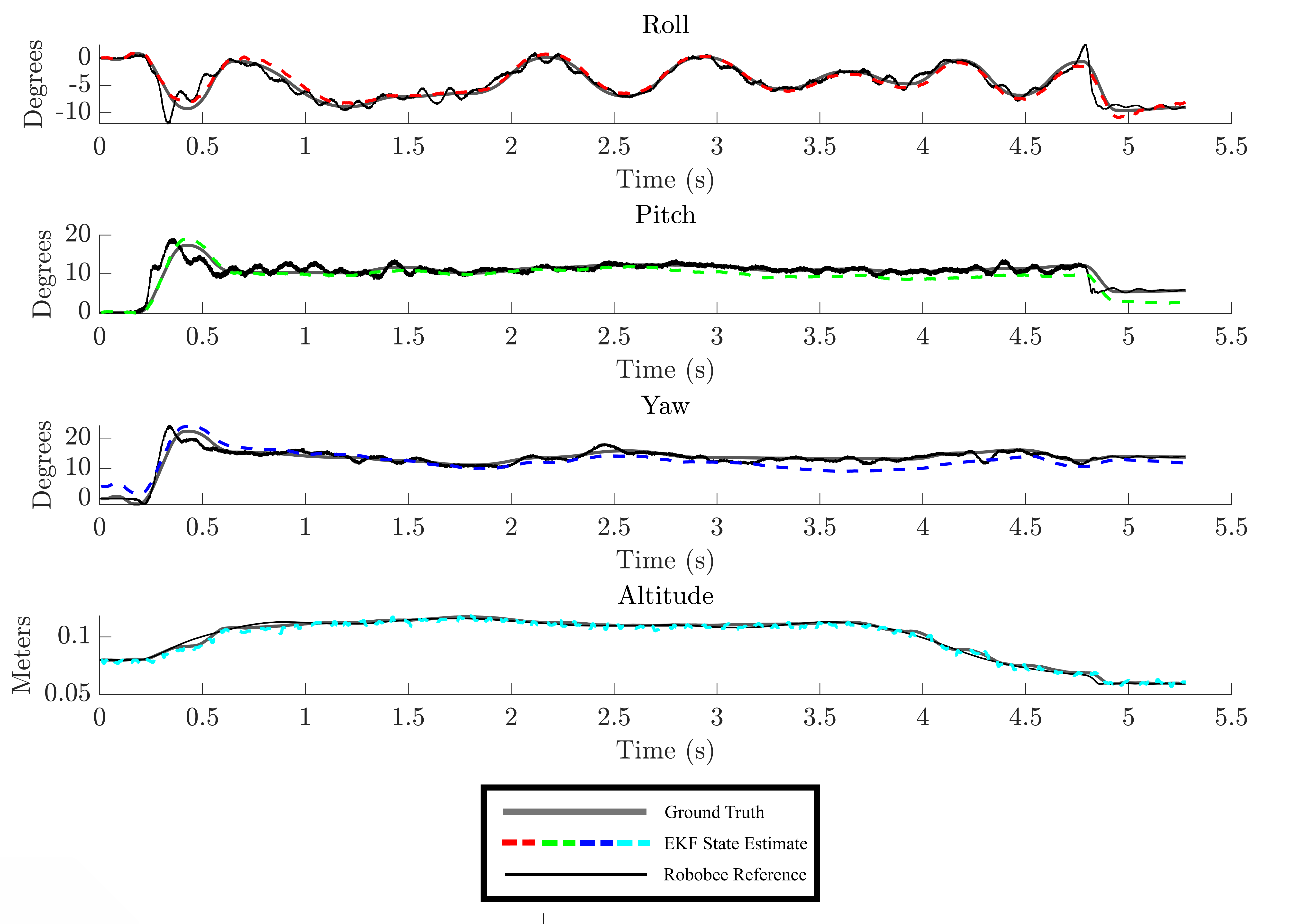}
    \caption{Vicon data from a RoboBee leaf hopping closed-loop flight experiment (gray, solid lines), plotted with the corresponding low pass filtered commanded UR5e trajectory, and the estimated trajectory with vibration noise applied. We observe minimal drift with less than 3 degree RMSE for the roll, pitch, and yaw and sub 2 mm RMSE for the altitude estimates for a 5.5 second flight experiment.}
    \label{fig:hovering_ekf_result}
\end{figure}

The estimated orientation and altitude for the leaf hopping reference flight experiment is plotted in Figure \ref{fig:hovering_ekf_result}, which not only plots the estimated state at the target induced noise amplitude, but also the true UR5e state and the original RoboBee's trajectory data. We observe that, despite the low-frequency body oscillation noise generated by the vibration motor, the CEKF is able to achieve an RMSE of $0.6^\circ$ for roll, $1.5^\circ$ for pitch, $2.2^\circ$ for yaw, and $2.1~\mathrm{mm}$ in altitude. Given that Fuller et. al. showed that the RoboBee was able to perform hovering maneuvers even with an RMSE of  $6^\circ$ in roll \cite{Fuller_Helbling_Chirarattananon_Wood_2014}, we can conclude that the accuracy of the CEKF algorithm satisfies the required orientation accuracy for RoboBee closed-loop flights.

Furthermore, we find that the accuracy of the CEKF algorithm does not degrade with even larger body oscillation noise than that experienced during actual RoboBee flight experiments. Table \ref{tab:results} shows that even at a 50\% duty cycle PWM, which corresponds to approximately 15 $\text{m}/\text{s}^2$ peak-to-peak oscillatory noise, the CEKF algorithm is able to achieve less than 3 degree RMSE orientation error. We begin to observe degradation in accuracy when the amplitude of the induced acceleration noise exceeds 20 $\text{m}/\text{s}^2$, at a 60-70\% duty cycle PWM, where the CEKF observes noticeably more drift, with up to 7 degree RMSE as seen in Table \ref{tab:results}. This is likely the point where the complementary filter struggles to separate the low frequency oscillation mode from the commanded trajectory, yielding inaccurate sensor orientation estimates with increasing drift over time. However, given that the magnitude of the body oscillation mode observed during real RoboBee flight experiments falls within the range of noise magnitude within which the CEKF algorithm produces accurate estimation, we conclude that the selected hardware and CEKF algorithm are capable of estimating the RoboBee state.

\section{Algorithm Computational Cost Analysis} \label{sec: discussion}

RoboBee's limited payload capacity not only limits the weight and size of the potential sensor suite, but it also indirectly limits computation capability. Currently, RoboBee flight experiments rely on external computers to gather sensor data, determine the current state, and generate appropriate control signals, providing the freedom to design controllers independent of computation complexity. The Vicon Motion Capture System used to determine the vehicle state is able to accomplish vision based state estimation in $<$ 2 $\mathrm{ms}$, providing accurate ($<$ 0.2 $\mathrm{mm}$ error) position and orientation estimates at 500Hz. However, as RoboBee continues to progress towards untethered, autonomous flight, efforts are being made to enable onboard compute which will significantly constrain the computation available for sensing and control algorithms. Given that we expect our proposed CEKF to run in real-time ($\approx$ 250 Hz), we evaluate the ability of the designed CEKF to perform in constrained computation for RoboBee's eventual onboard compute electronics. 

We characterize the computational cost of the proposed CEKF algorithm based on two factors: (1) the number of floating/fixed point operations (FLOPs), and (2) the applied number representation. We use these metrics to determine whether this algorithm is compatible with the low-power, low-cost embedded processors in consideration for onboard RoboBee control. Because the desired microprocessors often lack dedicated FPUs, frequent, high-precision operations would be impermissible on such systems. The subsequent discussion thus uses the Arm Cortex M0 as an example processor that satisfies the Harvard RoboBee's power constraints, with power consumption down to 50$\mu\text{W / MHz}$. We profile the proposed algorithm based on this platform and draw conclusions about its compatibility.

\begin{figure}[ht]
    \centering
    \includegraphics[width=\columnwidth]{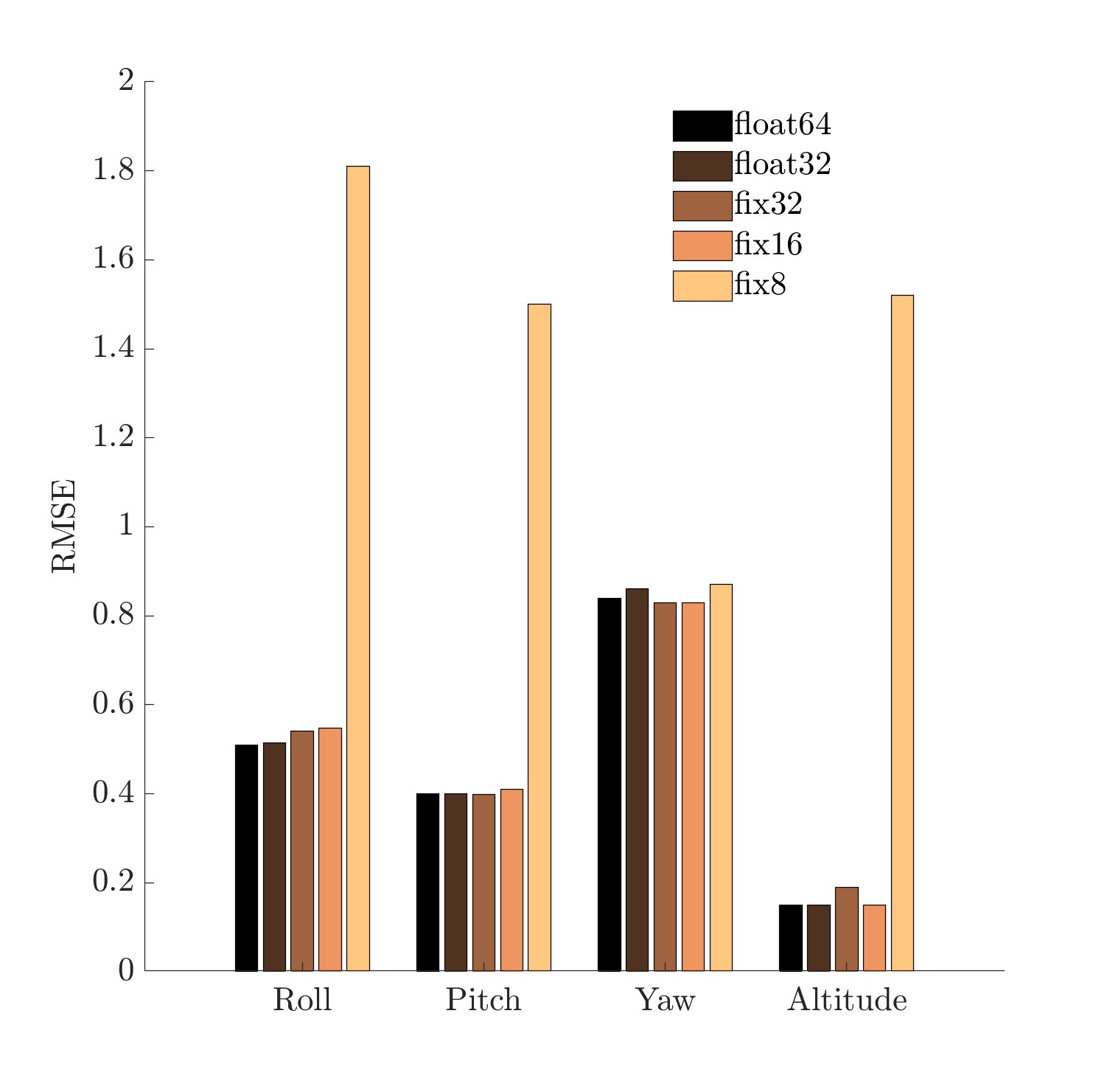}
    \caption{Bar plot of the RMSE accuracy, measured in degrees and cm, achieved by the CEKF for different number representations in memory from floating point 64 bit to fixed point 8 bit.}
    \label{fig:bit_acc}
\end{figure}

The proposed CEKF algorithm outlined in Section \ref{sec: CEKF} is dominated by FLOPS, requiring 861 multiplications and 225 additions. Although there are a few large matrix multiplications ($\mathbb{R}^{10 \times 10}$), the matrices are all sparse, with the major bottleneck being the computation of $(H\hat{P}^{(t+1)}H^T + R)^{-1}$ in Equation (\ref{eq: kalman gain}). Furthermore, given the memory constraints of embedded microcontrollers, such as the ARM Cortex M0, we profile the minimum required memory representation that does not compromise the performance substantially. Traditional computers operate with 64 bit floating point number representation which offers a numerical resolution of $4.9 \times 10^{-324}$. Meanwhile, 8 bit fixed point number representation only offers a resolution of $0.063$. Utilizing the collected sensor data from the leaf hopping hardware-in-the-loop experiment, with a 30\% duty cycle PWM noise generation, we compare the accuracy of the estimated orientation and altitude at different number representations and precisions. Figure \ref{fig:bit_acc} shows the resulting performance degradation as we move from a 64 bit floating point representation down to a fixed point 8 integer representation. We observe that with 16 bit fixed point representation, our CEKF algorithm achieves less than 1 degree and 1 mm RMSE for all state elements. Additionally, the significant loss in accuracy observed at a fixed point 8 bit representation in Figure \ref{fig:bit_acc} is a result of the EKF not having sufficient precision in parameters such as moments of inertia or mass to accurately predict the state.  We hypothesize that the reason the yaw RMSE does not increase as much as the others is because the original measurement estimate from the complementary filter is quite accurate and, consequently, the EKF's loss in precision does not compromise the yaw's estimate accuracy. Regardless, this property of the algorithm, which is likely an artifact of the convergence guarantees of EKF and the novel architecture of the CCF, is important as fixed point 16 bit number representations will allow microcontrollers such as the ARM Cortex M0 to perform 1063 FLOPS in $10.63$ microseconds, running at 100 MHz. The proposed algorithm comfortably satisfies our 250 Hz real-time constraint with ample slack, providing designers with the flexibility to tune their hardware and speed to meet power constraints.

\section{Conclusion}
 The physical scale and hardware interdependencies of FWMAVs makes the design and evaluation of potential sensor hardware and algorithms tedious. In this paper, we utilized a hardware in the loop approach to ``replay'' RoboBee trajectories in a rapid and reproducible manner on a UR5 arm as well as in a simulated environment from which we can extract sensor observations. We showed that the designed CEKF provides robust orientation and altitude estimates ($<$ 2 degrees, $<2$ mm) using sensors that satisfy the payload constraints for the current RoboBee design. Additionally, we showed that the algorithm was designed with onboard computing platforms in mind and still provides useful estimations using fixed-point representations down to 16 bits. Although the induced noise does not fully recreate RoboBee dynamics perfectly, our results provide evidence to support the algorithm and hardware performance and compatibility with a flight-worthy embedded perception system.

Having successfully demonstrated that our proposed algorithm and sensor suite meet the hardware and computation challenges of the Harvard RoboBee, the next steps involve bringing the sensing package onboard and integrating the estimated state within the flight controller. This endeavor necessitates a redesign of the RoboBee to integrate our sensor package (i.e., at the appropriate orientation and location relative to the vehicle center of mass) and re-tuning of control parameters to accommodate the additional payload and inertia. However, given our HWIL pipeline, we are confident that the deployment of this sensing package and state estimation algorithm onboard the RoboBee during flight experiments will be a critical step towards full autonomy.

\bibliographystyle{SageH}
\bibliography{bibtex/Bibliography}

\end{document}